\documentclass[conference]{IEEEtran}
\IEEEoverridecommandlockouts
\usepackage{cite}
\usepackage{amsmath,amssymb,amsfonts}
\usepackage{algorithmic}
\usepackage{graphicx}
\usepackage{textcomp}
\usepackage{xcolor}
\usepackage{booktabs}
\usepackage{multirow}
\usepackage{array}
\usepackage{dblfloatfix} 
\def\BibTeX{{\rm B\kern-.05em{\sc i\kern-.025em b}\kern-.08em
    T\kern-.1667em\lower.7ex\hbox{E}\kern-.125emX}}

\newcommand{\specialcell}[2][t]{\begin{tabular}[#1]{@{}l@{}}#2\end{tabular}}

\begin{document}

\title{Integrated Multi-omics Analysis Using Variational Autoencoders: Application to Pan-cancer Classification
\thanks{This project has received funding from the European Union’s Horizon 2020 research and innovation programme under the Marie Skłodowska-Curie grant agreement 764281.}
}

\author{\IEEEauthorblockN{Xiaoyu Zhang\textsuperscript{1}, Jingqing Zhang\textsuperscript{1}, Kai Sun\textsuperscript{1}, Xian Yang\textsuperscript{1}, Chengliang Dai\textsuperscript{1}, Yike Guo\textsuperscript{1,*}}
\IEEEauthorblockA{\textsuperscript{1}Data Science Institute, Imperial College London, London, SW7 2AZ, UK\\
\textsuperscript{*}Corresponding author, Email: y.guo@imperial.ac.uk}
}

\maketitle

\begin{abstract}
Different aspects of a clinical sample can be revealed by multiple types of omics data. Integrated analysis of multi-omics data provides a comprehensive view of patients, which has the potential to facilitate more accurate clinical decision making. However, omics data are normally high dimensional with large number of molecular features and relatively small number of available samples with clinical labels. The ``dimensionality curse'' makes it challenging to train a machine learning model using high dimensional omics data like DNA methylation and gene expression profiles. Here we propose an end-to-end deep learning model called OmiVAE to extract low dimensional features and classify samples from multi-omics data. OmiVAE combines the basic structure of variational autoencoders with a classification network to achieve task-oriented feature extraction and multi-class classification. The training procedure of OmiVAE is comprised of an unsupervised phase without the classifier and a supervised phase with the classifier. During the unsupervised phase, a hierarchical cluster structure of samples can be automatically formed without the need for labels. And in the supervised phase, OmiVAE achieved an average classification accuracy of 97.49\% after 10-fold cross-validation among 33 tumour types and normal samples, which shows better performance than other existing methods. The OmiVAE model learned from multi-omics data outperformed that using only one type of omics data, which indicates that the complementary information from different omics datatypes provides useful insights for biomedical tasks like cancer classification.
\end{abstract}


\section{Introduction}

The term ``omics'' refers to a discipline of life sciences analysing the interactions and functions of various biological entities in fields ending in the suffix -omics, encompassing genomics, epigenomics, transcriptomics, proteomics, metabolomics, etc. Thanks to the significant advances in high-throughput experimental technologies, a large amount of omics data in different datatypes is generated on an unprecedented scale \cite{berger2013computational}. Various types of omics data reveal different aspects of the same clinical samples. Integrated analysis of multi-omics data provides a comprehensive view of patients, which has the potential to facilitate more accurate clinical decision making.

Omics data, especially DNA methylation and gene expression profiles, are normally high dimensional with tens of thousands of or even hundreds of thousands of molecular features. However, the number of available samples with clinical labels like diagnostic and prognostic information is relatively small. To overcome the challenge of ``dimensionality curse'' and alleviate overfitting in task like cancer classification, routine methods manually select a small subset of the molecular features based on domain knowledge \cite{prat2012pam50} or use traditional dimensionality reduction algorithms like principal component analysis (PCA) before the downstream analysis, which may overlook some genome-wide hidden patterns. During the past few years, deep learning \cite{lecun2015deep} methods have shown great success in analysing high-dimensional data like images. Among them, variational autoencoders (VAE) \cite{kingma2013auto} have been widely applied to embed image and text data into lower dimensional latent spaces, whereas the application of VAE for analysing omics data is still in its early stage. 

In this work, we propose a model called OmiVAE for both low dimensional latent space extraction and multi-class classification on multi-omics datasets. OmiVAE combines a basic structure of a VAE with a classification network to achieve task-oriented feature extraction and end-to-end classification. We integrated genome-wide DNA methylation and gene expression profiles together with 450,804 molecular features from corresponding samples through this model. The training procedure of OmiVAE is comprised of an unsupervised phase without the classifier and a supervised phase with the classifier. Pan-cancer multi-omics datasets from The Cancer Genome Atlas (TCGA) \cite{weinstein2013cancer} with 9,081 samples of 33 tumour types and normal ones were used to evaluate this model. During the unsupervised phase, a hierarchical cluster structure of samples can be automatically formed without the need for labels. The higher-level clusters refer to different organs or systems. For a specific tissue-of-origin, various tumour subtypes and normal samples are grouped into corresponding lower-level clusters. In the supervised phase, OmiVAE achieved an average classification accuracy of 97.49\% using multi-omics data and 96.42\% using only gene expression data after 10-fold cross-validation among 33 tumour types and normal samples, which shows better performance than other existing methods. The OmiVAE model learned from multi-omics data outperformed that using only one type of omics data, which indicates that the complementary information from different omics datatypes provides useful insights for biomedical tasks like cancer classification.

\section{Related Work}

Inspired by the breakthrough in computer vision and natural language processing, many traditional machine learning and deep learning methods have been applied to multi-omics data analysis and cancer classification. For example, Chaudhary \textit{et al.} \cite{chaudhary2018deep} applied a traditional autoencoder (AE) framework to embed selected multi-omics molecular features (35,877 in total) into a 100-dimensional space and used learned features to predict survival-risk subgroups of liver cancer patients by \textit{k}-means clustering and support vector machine (SVM). Chung \textit{et al.} \cite{chung2019unsupervised} proposed a VAE model based on long short-term memory (LSTM) for cardiac remodeling using metabolomics and proteomics time-series data. The models built by these studies were not end-to-end and the features learned by AE/VAE were used for downstream tasks outside the network. Ma and Zhang \cite{ma2018multi} proposed an end-to-end multi-omics model using factorization autoencoder to predict disease progression-free interval events with an average precision of 0.664 for bladder tumour dataset and 0.746 for glioma dataset.

For cancer classification, Fakoor \textit{et al.} \cite{fakoor2013using} proposed a model that combined PCA and sparse autoencoder to learn a representation of the original gene expression data and used the learned features for cancer classification. Stacked denoising autoencoder (SDAE), another variation of autoencoder, was also applied to extract features from gene expression profiles \cite{danaee2017deep}. Way and Greene \cite{way2018extracting} used a VAE model to extract a latent space from gene expression data and analysed the relationship between latent variables and certain phenotypes. Other methods \cite{titus2018new, wang2018exploring} also applied VAE to reduce the dimension of methylation data and used the latent variables to classify subtypes of lung and breast tumours. Rhee \textit{et al.} \cite{ijcai2018-490} proposed a hybrid model integrating graph convolution neural network (CNN) and relation network to classify breast cancer subtypes using gene expression profiles and protein-protein interaction (PPI) networks. In a more comprehensive study, Capper \textit{et al.} \cite{capper2018dna} developed a central nervous system (CNS) tumours classification system based on DNA methylation data using a random forest algorithm and evaluated this method in actual clinical implementation with promising results. 

Most of the above work only focused on cancer identification or classification among certain types of cancers. As for pan-cancer classification, Li \textit{et al.} \cite{li2017comprehensive} trained a model combined a genetic algorithm and \textit{k}-nearest neighbors (KNN) on the TCGA gene expression datasets and achieved an overall prediction accuracy of 95.6\%. Lyu and Haque \cite{lyu2018deep} reshaped gene expression profiles into 2D images and applied a CNN to classify 33 tumour types from the TCGA datasets. The average accuracy of this model was 95.59\%. Mostavi \textit{et al.} \cite{mostavi2019convolutional} implemented three CNN models on the same datasets and achieved 95.7\% accuracy for 33 tumour types and 95.0\% accuracy for tumour types and normal samples. All of these studies only used gene expression data for pan-cancer classification without including complementary information from other types of omics data.

\section{Materials and Methods}

\subsection{Datasets}
Two types of high dimensional omics data, RNA-Seq gene expression profiles and DNA methylation profiles were used in OmiVAE. We chose TCGA pan-cancer datasets with 33 various tumour types for both gene expression and DNA methylation data. The gene expression dataset consists of 11,538 samples, among which 741 samples are normal tissues. Each gene expression profile is comprised of 60,483 identifiers referring to corresponding exons. Fragments Per Kilobase of transcript per Million mapped reads (FPKM) values in this dataset were $\log_{2}$-transformed. The DNA methylation dataset contains 9,736 samples, in which 746 samples are normal tissues. DNA methylation profiles in this dataset were produced by Infinium HumanMethylation450 BeadChip (450K) arrays with 485,578 probes. The Beta value of each probe indicates the methylation ratio of the corresponding CpG site. Both datasets were downloaded from UCSC Xena data portal\footnote{https://xenabrowser.net/datapages/} on April 16th, 2019.

\subsection{Data Preprocessing}

Data preprocessing procedures were applied to the datasets to remove noises and biassed signals. We filtered molecular features (\textit{i.e.,} exons or CpG sites) according to the following criteria. For gene expression data, exons located at the Y chromosome (n = 594),  exons with zero expression level in all samples (n = 1,904) and exons that have missing values (N/A) in more than 10\% of samples (n = 248) were removed. In total, 2,440 molecular features were removed and 58,043 were kept in the gene expression data. 

As for DNA methylation data, probes that can not be mapped to the human reference genome (hg38) annotation (n = 89,512), control probes that are not located at any chromosome (n = 2,545), probes targeting the Y chromosome (n = 346) and probes with missing values in more than 10\% samples (n = 414) were filtered out, which results in a final DNA methylation feature set of 392,761 CpG sites. These probes were then grouped into 23 subsets according to their targeting chromosomes. Different chromosomes have different numbers of probes and the average is 17,077.

After the filtering step, the remaining missing values in both datasets were replaced by the mean of corresponding molecular features. For RNA-Seq gene expression data, the $\log_{2}$-transformed FPKM measures were normalised to the range of 0 to 1 due to the input requirement of our model. This normalisation step is not necessary for Beat values in the DNA methylation dataset, because they are defined in percentage. Samples in TCGA that have both gene expression and DNA methylation profiles were used for multi-omics analysis. We collected 9081 samples in total, in which 407 were normal tissue samples.

\subsection{VAE/Classification Network}

Variational autoencoder (VAE) \cite{kingma2013auto} is a powerful deep generative model capable of learning meaningful data manifold from high dimensional data. Given a multi-omics dataset $\mathcal{D}$ that is comprised of $N$ clinical samples $\left\{\mathbf{x}^{(i)}\right\}_{i=1}^{N}$ with $d$ multi-omics molecular features, we assume each sample $\mathbf{x}^{(i)} \in \mathbb{R}^d$ is generated by a latent vector $\mathbf{z}^{(i)} \in \mathbb{R}^p$, where $p \ll d$. This generation process comprises two major steps: each latent variable $\mathbf{z}^{(i)}$ is generated from prior distribution $p_\theta(\mathbf{z})$, and each sample $\mathbf{x}^{(i)}$ is generated from conditional distribution $p_\theta(\mathbf{x|z})$, where $\theta$ is the set of learnable parameters of the generative network (or decoder). Thereby, the objective of a generative model with latent variables is the distribution of $\mathbf{x}$, which is:
\begin{equation}
    p_{\theta}(\mathbf{x})=\int_{z} p_{\theta}(\mathbf{x|z}) p_{\theta}(\mathbf{z}) d z
    \label{eq:generative_model}
\end{equation}
The integral in (\ref{eq:generative_model}) is computationally infeasible, due to the intractability of the true posterior $p_\theta(\mathbf{z|x})$. In order to address this issue, variational distribution $q_{\phi}(\mathbf{z|x})$ is introduced to approximate $p_\theta(\mathbf{z|x})$, where $\phi$ is the set of learnable parameters of the inference network (or encoder). In the VAE framework, the encoder and the decoder are jointly optimised by maximizing the variational lower bound:
\begin{equation}
    \mathbb{E}_{\mathbf{z} \sim q_{\phi}(\mathbf{z|x})} \log p_{\theta}(\mathbf{x|z}) - D_{\mathrm{KL}}\left(q_{\phi}(\mathbf{z|x}) \| p_\theta(\mathbf{z})\right)
\end{equation}
where $D_{\mathrm{KL}}$ is the Kullback–Leibler (KL) divergence between two probability distributions \cite{goodfellow2016deep}.

\begin{figure}[tb]
    \centering
    \includegraphics[scale=0.83]{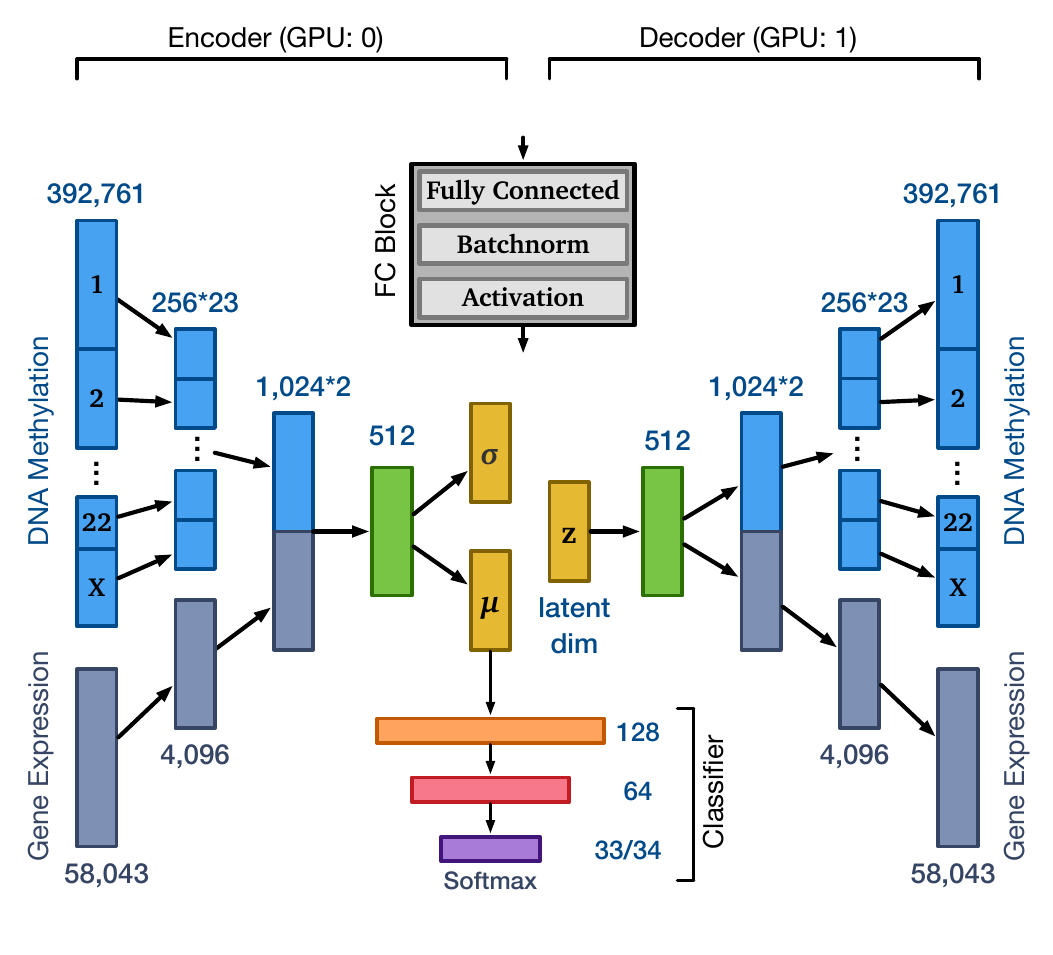}
    \caption{Diagram showing the architecture of OmiVAE. The network structure is comprised of three main components: an encoder, a decoder, and a classifier. Each rectangle denotes a fully connected (FC) block, the detailed structure of which is shown in the gray box. FC blocks for gene expression and DNA methylation data are marked in different colours. CpG sites in methylation profiles are separated into different blocks according to their targeting chromosomes. Model parallelism is applied by assigning the encoder to GPU:0 and assigning decoder to GPU:1.}
    \label{fig:network_structure}
\end{figure}

In OmiVAE, we combines the basic structure of VAE with a classification network to achieve task-oriented feature extraction and end-to-end cancer classification. The architecture of OmiVAE is illustrated in Fig. \ref{fig:network_structure}. The network structure is comprised of three main components: an encoder, a decoder, and a classifier. 

\begin{figure*}[b]
    \centering
    \includegraphics[scale=0.59]{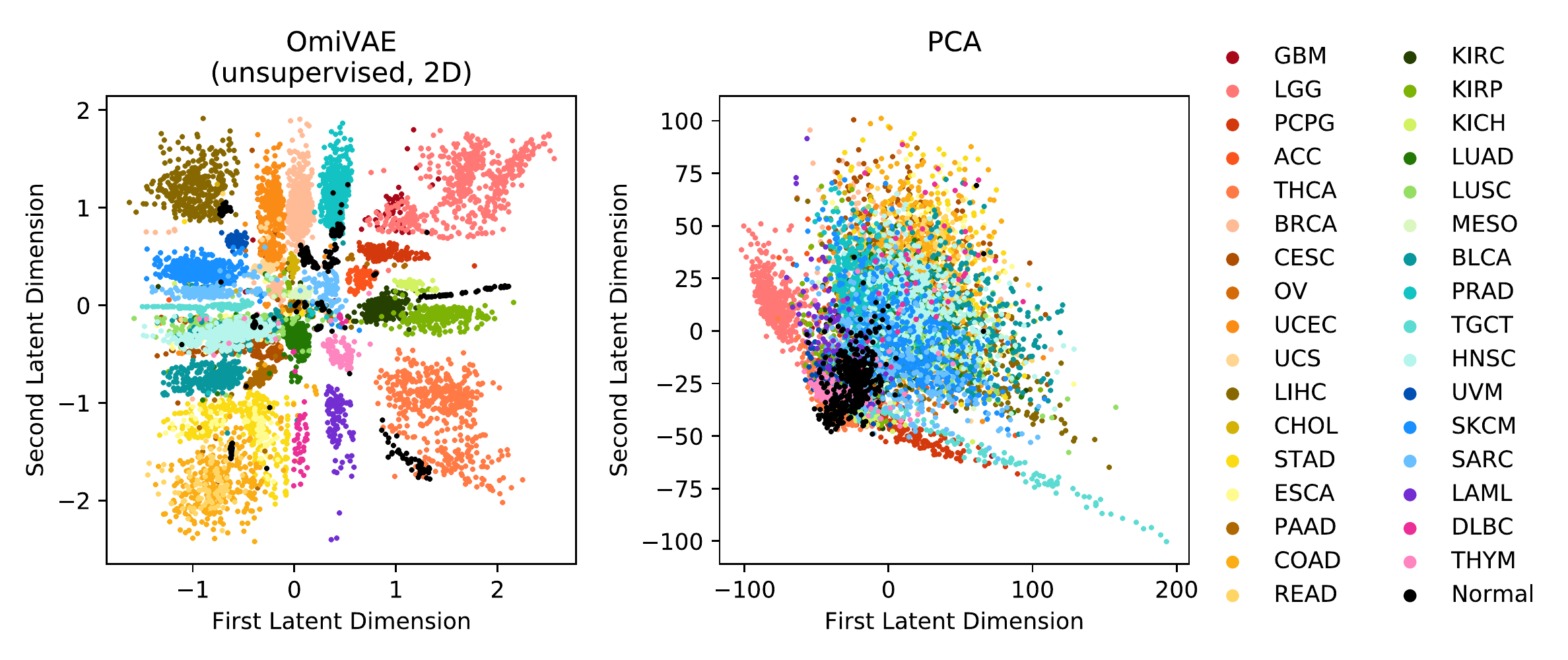}
    \caption{2D embedding of the multi-omics data learned by the unsupervised phase of OmiVAE (left) and PCA (right). Samples of various tumour types were plotted with different colours shown in the legend. Normal samples were marked in black dots. The full names of the tumour type abbreviations can be found on https://gdc.cancer.gov/resources-tcga-users/tcga-code-tables/tcga-study-abbreviations.}
    \label{fig:compare}
\end{figure*}

In the encoder network, gene expression and DNA methylation profiles are first encoded into two 1,024 dimensional vectors separately through two hidden layers. After data preprocessing, in each methylation profile there are still 392,761 molecular features, which is almost ten times larger than the feature size of a gene expression profile. These methylation features are separated into different fully connected blocks according to their targeting chromosomes in order to reduce the number of parameters in fully connected layers that encode methylation data, prevent overfitting and save GPU memory for training. Features in each chromosome are encoded into corresponding vectors with 256 dimensions in the first hidden layer to capture the intra-chromosome relationships, whereas inter-chromosome relationships are obtained through the second hidden layer. The two 1,024 dimensional vectors are concatenated together in the second hidden layer and encoded to a multi-omics vector. The final hidden layer of the encoder fully connects to two output layers, which represent the mean $\boldsymbol{\mu}$ and the standard deviation $\boldsymbol{\sigma}$ in the Gaussian distribution $\mathcal{N}\left(\boldsymbol{\mu}, \boldsymbol{\sigma}\right)$ of the latent variable $\mathbf{z}$ given input sample $\mathbf{x}$, which is the variational distribution $q_{\phi}(\mathbf{z|x})$. In order to make the sampling step differentiable and suitable for backpropagation, the reparameterization trick is applied as follows:
\begin{equation}
    \mathbf{z}=\boldsymbol{\mu}+\boldsymbol{\sigma} \boldsymbol{\epsilon}
\end{equation}
where $\boldsymbol{\epsilon}$ is a random variable sampled from the unit normal distribution $\mathcal{N}(\mathbf{0}, \mathbf{I})$. The dimension of the bottleneck layer (\textit{i.e.,} $\boldsymbol{\mu}$, $\boldsymbol{\sigma}$, and $\mathbf{z}$) is normally set to 128. The outputs of the bottleneck layer can also be used to visualise the clusters of samples, and for this purpose the dimension of the bottleneck layer can be set to 2 or 3 directly for 2D or 3D scatter plots.

The network structure of the decoder is similar to the mirror image of the encoder network. Latent variable $\mathbf{z}$ is the input of the decoder network followed by three hidden layers. The output of the decoder is the reconstructed vector $\mathbf{x}^\prime$ which consists of both the DNA methylation and gene expression profile. As for the classifier, the output vector $\boldsymbol{\mu}$ in the encoder is connected to a three-layer classification network to predict whether $\mathbf{x}$ is a tumour sample and the tumour type of this sample. The output dimension of the classifier is set to 33 for the tumour type classification and set to 34 for classification of both the 33 tumour types and normal samples. This classifier part adds further regularisation to the lower dimensional latent representation with the classification task. In a basic VAE model, the bottleneck layer tends to extract the most essential features that can reconstruct input samples as closely as possible. Nevertheless these extracted features may not be related to specific task like tumour type classification. With the regularisation of the classifier, the network is encouraged to learn latent representations that can not only accurately reconstruct the input sample but also identify cancer and classify types. 

Base on the aforementioned idea, the joint loss function of the whole network can be defined as the combination of a VAE loss and a classification loss. We denote the gene expression profile of the input sample $\mathbf{x}$ as $\mathbf{x}_{e}$ and the reconstructed gene expression vector as $\mathbf{x}_{e}^{\prime}$. For DNA methylation profile, molecular features are grouped into 23 input vectors according to their targeting chromosomes. These vectors are denoted as $\mathbf{x}_{m_j}$ where $j$ is the index of the corresponding chromosome. The VAE loss can then be written as follows:
\begin{equation}
    \mathcal{L}_{vae}=\frac{1}{M} \sum_{j=1}^{M} CE \left(\mathbf{x}_{m_j}, \mathbf{x}_{m_j}^{\prime} \right)+ CE\left(\mathbf{x}_{e}, \mathbf{x}_{e}^{\prime}\right) + \mathcal{L}_{KL}
\end{equation}
where $M$ is the number of chromosomes (23 in our case), $CE$ is the binary cross-entropy between input vectors and reconstructed vectors, $\mathcal{L}_{KL}$ is the KL divergence between the learned distribution and a unit Gaussian distribution, which is:
\begin{equation}
    \mathcal{L}_{KL} = D_{\mathrm{KL}}(\mathcal{N}(\boldsymbol{\mu}, \boldsymbol{\sigma}) \| \mathcal{N}(\mathbf{0}, \mathbf{I})).
\end{equation}
For the classifier, the ground truth label of a given sample $\mathbf{x}$ is denoted as $y$, and the label predicted by the classifier is denoted as $y^{\prime}$. The classification loss can be defined as,
\begin{equation}
    \mathcal{L}_{class} = CE(y,y^{\prime})
\end{equation}
where $CE$ is the cross-entropy loss between the ground truth and the predicted label. The total loss function of OmiVAE is
\begin{equation}
    \mathcal{L}_{total} = \alpha \mathcal{L}_{vae} + \beta \mathcal{L}_{class}
    \label{eq:total_loss}
\end{equation}
and depends on $\alpha$ and $\beta$ parameters which weight the two main terms during training.

As for the model implementation, batch normalisation \cite{ioffe2015batch} and certain activation functions are added in each fully connected block. For most hidden layers, rectified linear units (ReLU) \cite{nair2010rectified} is selected as the activation function, whereas sigmoid activation is applied to the output layer of the decoder and softmax is attached to the final layer of the classifier. The model was built in PyTorch (version 1.1) and trained on two NVIDIA GeForce GTX 1080 Ti GPUs with 11 gigabytes of memory each. In order to take full advantage of the two GPUs and separate the enormous number of parameters ($\sim 7 \times 10^8$) into them, we applied a model parallelism strategy that assigned the encoder to the first GPU and assigned the decoder to the second GPU. The implementation of our proposed model has been made publicly available on GitHub\footnote{https://github.com/zhangxiaoyu11/OmiVAE}.

\section{Results}

We demonstrated the performance of our model on the aforementioned preprocessed multi-omics datasets which were randomly divided into training, validation and testing sets, proportionally allocating samples from each class. The network architecture was implemented as same as the diagram shown in Fig. \ref{fig:network_structure}. Adam optimiser \cite{kingma2015adam} was utilised in this model with the learning rate equaled to $10^{-3}$ and a batch size of 32. The early stopping technique was applied to avoid overfitting and we used stratified 10-fold cross-validation to robustly evaluate the performance of OmiVAE and other methods to avoid the bias from a single testing set. The training procedure of OmiVAE was comprised of an unsupervised phase without the classifier and a supervised phase with the classifier. In the unsupervised phase, the weight $\beta$ in (\ref{eq:total_loss}) was set to 0 and no label information was used at this stage. OmiVAE can be regarded as an unsupervised feature extraction and dimensionality reduction model if only this phase is applied. In the supervised phase, the weight $\beta$ was set to 1 to extract features that were related to the pan-cancer classification task and classify samples in an end-to-end manner.

\begin{figure}[tb]
    \centering
    \includegraphics[scale=0.58]{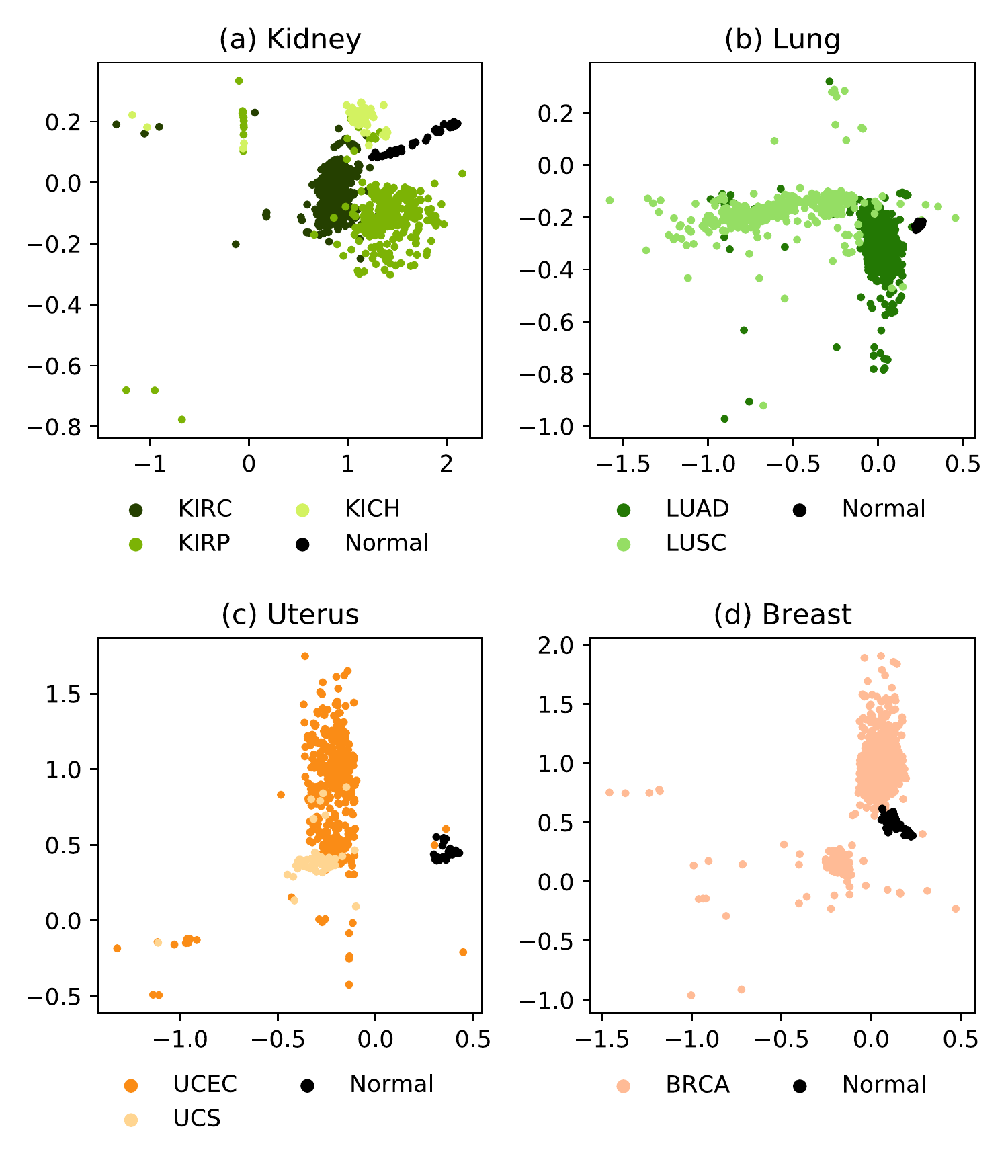}
    \caption{Multi-omics latent vectors of tumours and corresponding normal samples from (a) kidney, (b) lung, (c) uterus and (d) breast learned by the unsupervised phase of OmiVAE.}
    \label{fig:organ}
\end{figure}

\subsection{Unsupervised Phase}

In our experiments, we first set the latent dimension directly to 2 so as to visualise the latent space and evaluate the dimension reduction ability of OmiVAE. The two-dimensional embedding of the 450,804 multi-omics features for all 9,081 samples learned by OmiVAE were plotted on a scatter graph shown in the left-side of Fig. \ref{fig:compare}. As shown in the scatter graph, a hierarchical cluster structure of input samples was automatically formed through the latent space embedding of OmiVAE. The higher-level clusters represent different organs or systems. For a specific tissue-of-origin, samples of various tumour subtypes and normal samples were grouped into corresponding lower-level clusters. These results indicate that the low dimensional manifold learned by OmiVAE preserves both the global and local clustering structures of the input samples.

In order to further evaluate the performance of the low dimensional embedding and delve into the two-level hierarchical cluster structure of the latent space, we separately checked samples from specific organs with different tumour subtypes and normal samples. In the scatter graph shown in Fig. \ref{fig:organ} (a), we selected all kidney samples in the TCGA datasets from patients with three subtypes of kidney tumours, kidney clear cell carcinoma (KIRC), kidney papillary cell carcinoma (KIRP) and kidney chromophobe (KICH). Normal samples from three different cohorts were mapped to the same cluster through the 2-dimensional embedding of OmiVAE, and the three subtypes of kidney tumours were also mapped to three distinct clusters separately. We also investigated tumour types and normal samples from lung, uterus, and breast, results of which were shown in Fig. \ref{fig:organ} (b-d). In all of these scatter graphs, normal samples can be easily separated from tumour ones in the latent space. For lung and uterus, both of them have two tumour subtypes which can be clearly distinguished from each other in the low dimensional representation. The results shown in Fig. \ref{fig:organ} indicate that the representations of high-dimensional multi-omics data learned by the unsupervised phase of OmiVAE are discriminative for the downstream classification tasks of tumour types and normal samples. 

\begin{table}[tb]
    \caption{Classification Performance of Representations Learned by Different Dimensionality Reduction Methods}
    \centering
    \begin{tabular}{l|c|c|c|c}
        \toprule
        & Accuracy & Precision & Recall & F1 Score \\  \midrule[0.8pt]
        PCA+SVM & 30.13$\pm$1.62\% & 0.26$\pm$0.02 & 0.30$\pm$0.02 & 0.26$\pm$0.02 \\ 
        \midrule
        KPCA+SVM & 30.16$\pm$1.65\% & 0.26$\pm$0.02 & 0.30$\pm$0.02 & 0.26$\pm$0.02 \\ 
        \midrule
        t-SNE+SVM & 82.94$\pm$0.87\% & 0.80$\pm$0.01 & 0.83$\pm$0.01 & 0.80$\pm$0.01 \\ 
        \midrule
        UMAP+SVM & 80.39$\pm$0.96\% & 0.73$\pm$0.01 & 0.80$\pm$0.01 & 0.76$\pm$0.01 \\ 
        \midrule
        \specialcell{OmiVAE+SVM \\ (1st phase, 2D)} & \multirow{2}{*}{\textbf{84.40$\pm$0.75\%}} & \multirow{2}{*}{\textbf{0.83$\pm$0.01}} & \multirow{2}{*}{\textbf{0.84$\pm$0.01}} & \multirow{2}{*}{\textbf{0.82$\pm$0.01}} \\ \bottomrule
    \end{tabular}
    \label{tab1}
\end{table}

The performance of the unsupervised phase of OmiVAE was compared with other feature extraction and dimensionality reduction methods including the most widely-used algorithm PCA, its non-linear variant KPCA, the popular manifold learning method t-distributed stochastic neighbor embedding (t-SNE) \cite{maaten2008visualizing} and the latest dimension reduction technique uniform manifold approximation and projection (UMAP) \cite{mcinnes2018umap}. We first applied PCA to embed the same multi-omics dataset to a 2D space. The low dimensional representations learned by PCA were plotted on the scatter graph shown in the right-side of Fig. \ref{fig:compare}. As can be seen in the two graphs, the 2D embedding learned by PCA shows a much more mixed clustering structure compared with that learned by the unsupervised phase of OmiVAE. Samples of different tumour types are very difficult to discriminate from each other. In order to evaluate the feature extraction ability of OmiVAE and other existing methods in a more quantified manner, we fed the 2D representations learned by these methods to a support vector machine (SVM) with a radial basis function (RBF) kernel and used the SVM to classify tumour and normal samples (34 classes in total). 

The classification performance of the five methods was measured by four multi-class classification metrics: overall accuracy, weighted precision, weight recall, and weighted F1 score. The performance results after 10-fold cross-validation are shown in Table \ref{tab1}. As we can see in the table, the 2D representations learned by the unsupervised phase of OmiVAE outperformed those learned by other dimensionality reduction methods in all the four classification metrics. Note that the metrics standard deviations of OmiVAE are also the least, which indicates that the performance of OmiVAE is more robust compared with other methods. The method with closest performance to the unsupervised phase of OmiVAE is t-SNE. In addition to the less discriminative extracted features, there are also other drawbacks of t-SNE compared with OmiVAE. First, unlike OmiVAE, t-SNE is not a parametric method that learns a mapping function which can be used for new data. For data outside the training set, t-SNE needs to combine the data together and rerun the whole process. Second, although t-SNE is able to adequately keep the local clustering structures, the global structure is not explicitly preserved. Finally, t-SNE method can only embed data into two or three dimensional space, while the dimension of the bottleneck layer of OmiVAE can be set to any number.

\subsection{Supervised Phase}
In the supervised phase, the weight $\beta$ in (\ref{eq:total_loss}) was set to 1 and labels were used to train the classifier. For the classification task, the latent dimension was set to 128 to preserve more information from the high-dimensional multi-omics data during the whole training process including the unsupervised phase, which can be considered as a pre-training step in OmiVAE. Parameters of the encoder and the decoder learned in the unsupervised phase were transferred to the supervised phase and can still be fine-tuned to extract features more related to the classification task during the second phase.

\begin{table}[tb]
    \caption{Classification Performance of OmiVAE on Different Tumour Types and Normal Samples (34 classes)}
    \centering
    \begin{tabular}{l|l|c|c}
        \toprule
            & Application Mode & Accuracy & F1 Score \\ \midrule[0.8pt]
        \multirow{4}{*}{\specialcell{Only Gene \\ Expression }} & \specialcell{OmiVAE+SVM \\ (unsupervised phase)} & \multirow{2}{*}{93.12$\pm$0.54\%} & \multirow{2}{*}{0.926$\pm$0.006} \\ \cmidrule{2-4}
            & \specialcell{OmiVAE \\ (end-to-end model)} & \multirow{2}{*}{\textbf{96.37$\pm$0.46\%}} & \multirow{2}{*}{0.963$\pm$0.005} \\ \midrule
        \multirow{4}{*}{\specialcell{Only DNA \\ Methylation }} & \specialcell{OmiVAE+SVM \\ (unsupervised phase)} & \multirow{2}{*}{91.10$\pm$0.92\%} & \multirow{2}{*}{0.899$\pm$0.010} \\ \cmidrule{2-4}
            & \specialcell{OmiVAE \\ (end-to-end model)} & \multirow{2}{*}{\textbf{96.37$\pm$0.70\%}} & \multirow{2}{*}{0.963$\pm$0.007} \\ \midrule
        \multirow{4}{*}{\specialcell{Multi-Omics \\ Data }} & \specialcell{OmiVAE+SVM \\ (unsupervised phase)} & \multirow{2}{*}{94.16$\pm$0.74\%} & \multirow{2}{*}{0.937$\pm$0.008} \\ \cmidrule{2-4} 
            & \specialcell{OmiVAE \\ (end-to-end model)} & \multirow{2}{*}{\textbf{97.49$\pm$0.45\%}} & \multirow{2}{*}{0.975$\pm$0.005} \\ \bottomrule
    \end{tabular}
    \label{tab2}
\end{table}

We first evaluated the classification performance of OmiVAE on the TCGA multi-omics datasets with both the tumour samples and normal samples. There are 34 classes (33 tumour types plus normal samples) in this multi-class classification task. The results of OmiVAE on this task are shown in Table \ref{tab2}. Two application modes of OmiVAE were compared with each other in the experiments. As can be seen in the table, the classification performance of the end-to-end version of OmiVAE is better than that of the model where features learned by the unsupervised phase of OmiVAE were fed to an SVM for classification. This could be due to that some features extracted by the unsupervised phase of OmiVAE may not be specifically related to the cancer classification task.

In order to test whether the information from  different types of omics data is complementary and can therefore improve the performance of classification tasks, we applied the gene expression dataset and DNA methylation dataset separately to OmiVAE and evaluated the classification performance with only one type of omics data. As shown in Table \ref{tab2}, the results obtained from gene expression data and DNA methylation data are close to each other, while both of them are inferior to the performance of multi-omics data, which occurred in both application modes of OmiVAE (only the unsupervised phase and the end-to-end mode). This indicates that the complementary information from different omics datatypes is combined together in the latent vector learned by OmiVAE to extract more aspects of the input samples and can therefore be helpful for biomedical tasks like cancer classification.

\begin{figure}[tb]
    \centering
    \includegraphics[scale=0.52]{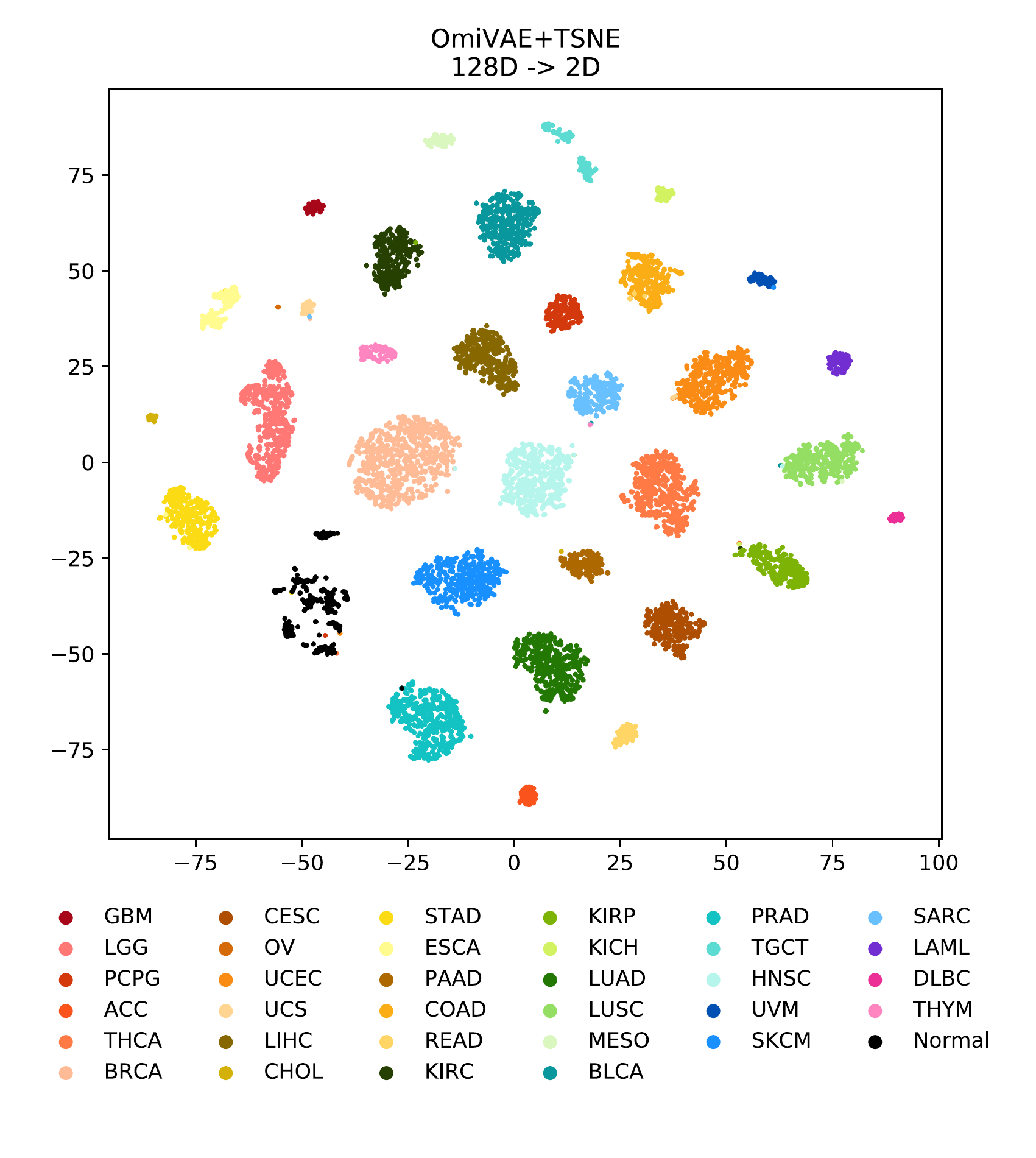}
    \caption{Visualisation of the classification space learned by the end-to-end OmiVAE. t-SNE was applied to reduce the dimensionality from 128D to 2D for scatter plotting.}
    \label{fig:omivae_tsne}
\end{figure}

The average classification accuracy on different tumour types and normal samples (34 classes) achieved by the end-to-end OmiVAE is 97.49\% after 10-fold cross-validation and the standard deviation is 0.45\%. As for the performance of other methods, only Mostavi \textit{et al.} \cite{mostavi2019convolutional} evaluated their model on the 34-class task for both tumour and normal samples with an accuracy of 95.0\%. To further visualise the classification space learned by OmiVAE, we used t-SNE to reduce the dimension of latent vectors from 128 to 2 and plotted samples on a 2D scatter graph shown in Fig. \ref{fig:omivae_tsne}. Clusters of normal samples and different types of tumour samples can be clearly recognised in this scatter graph.

We also evaluated our method on the tumour type classification task specifically by removing all normal samples in the TCGA multi-omics datasets. There are 33 classes corresponding to 33 tumour types in this pan-cancer classification task. The performance of OmiVAE on this task is shown in Table \ref{tab3}. The classification performance of both application modes on the 33-class task is slightly better than those on the 34-class task. The average accuracy achieved by the end-to-end OmiVAE is 97.88\%, which outperformed other existing methods \cite{li2017comprehensive, lyu2018deep, mostavi2019convolutional} on the same classification task. In the results, we can also observe that models trained on the multi-omics data achieved better performance than those trained on only one type of omics data, which we have mentioned earlier.

\begin{table}[tb]
    \caption{Classification Performance of OmiVAE Only on Different Tumour Types (33 classes)}
    \centering
    \begin{tabular}{l|l|c|c}
        \toprule
            & Application Mode & Accuracy & F1 Score \\ \midrule[0.8pt]
        \multirow{4}{*}{\specialcell{Only Gene \\ Expression }} & \specialcell{OmiVAE+SVM \\ (unsupervised phase)} & \multirow{2}{*}{93.66$\pm$0.57\%} & \multirow{2}{*}{0.931$\pm$0.006} \\ \cmidrule{2-4}
            & \specialcell{OmiVAE \\ (end-to-end model)} & \multirow{2}{*}{\textbf{96.86$\pm$0.48\%}} & \multirow{2}{*}{0.968$\pm$0.005} \\ \midrule
        \multirow{4}{*}{\specialcell{Only DNA \\ Methylation }} & \specialcell{OmiVAE+SVM \\ (unsupervised phase)} & \multirow{2}{*}{92.04$\pm$0.38\%} & \multirow{2}{*}{0.909$\pm$0.004} \\ \cmidrule{2-4}
            & \specialcell{OmiVAE \\ (end-to-end model)} & \multirow{2}{*}{\textbf{96.89$\pm$0.45\%}} & \multirow{2}{*}{0.968$\pm$0.005} \\ \midrule
        \multirow{4}{*}{\specialcell{Multi-Omics \\ Data }} & \specialcell{OmiVAE+SVM \\ (unsupervised phase)} & \multirow{2}{*}{94.51$\pm$0.48\%} & \multirow{2}{*}{0.940$\pm$0.005} \\ \cmidrule{2-4} 
            & \specialcell{OmiVAE \\ (end-to-end model)} & \multirow{2}{*}{\textbf{97.88$\pm$0.15\%}} & \multirow{2}{*}{0.979$\pm$0.002} \\ \bottomrule
    \end{tabular}
    \label{tab3}
\end{table}

\section{Conclusion}
Different types of omics data reveal various aspects of the same samples. Integrated analysis of multi-omics data allows us to get a comprehensive picture of patients. However, the number of available labeled samples and the dimensionality of multi-omics data are normally mismatched, which is called the ``dimensionality curse''. To address this problem, we proposed an end-to-end deep learning model called OmiVAE for both low dimensional latent space extraction and multi-class classification on multi-omics datasets. We integrated genome-wide DNA methylation and gene expression profiles together with more than 450,000 molecular features through this model. A hierarchical cluster structure of samples can be automatically formed without the need for labels during the unsupervised phase of this model. In the supervised phase, OmiVAE achieved an average classification accuracy of 97.88\% on different tumour types and 97.49\% on tumour types and normal samples after 10-fold cross-validation, which is better than existing methods. Moreover, the OmiVAE model learned from multi-omics data outperformed that using only one type of omics data, which indicates that the complementary information from different omics datatypes provides useful insights for biomedical tasks like cancer classification.

\bibliographystyle{IEEEtran}
\bibliography{bibm}

\end{document}